\documentclass{article} % For LaTeX2e
\usepackage{scoreclr_arkiv,times}
\usepackage{amsthm}
\newtheorem{proposition}{Proposition}[section]
\newtheorem{assumption}{Assumption}[section]
\usepackage{algorithm}
\usepackage{algpseudocode}
\usepackage{booktabs}
\usepackage{graphicx}

\usepackage{amsmath,amsfonts,bm}

\def\eqref#1{equation~\ref{#1}}
\def\1{\bm{1}}

\DeclareMathAlphabet{\mathsfit}{\encodingdefault}{\sfdefault}{m}{sl}
\SetMathAlphabet{\mathsfit}{bold}{\encodingdefault}{\sfdefault}{bx}{n}

\usepackage{hyperref}
\usepackage{url}

\title{Contrastive latent representation learning with denoising score matching}

\author{Christopher Marouani\thanks{Work done during an internship at the University of Amsterdam.} \\
ENS Paris-Saclay \\
University of Amsterdam \\
\texttt{christopher.marouani@ens-paris-saclay.fr} \\
\And
Grigory Bartosh \\
University of Amsterdam \\
\texttt{g.bartosh@uva.nl} \\
\And
Christian A. Naesseth \\
University of Amsterdam \\
\texttt{c.a.naesseth@uva.nl}
}

\iclrfinalcopy
\begin{document}

\maketitle

\begin{abstract}
Contrastive representation learning methods operate by pulling different views of the same data sample together while pushing them away from views of different samples. A fundamental challenge is the estimation of the repulsive force, which typically requires large batch sizes, can exhibit high variance, and is susceptible to the curse of dimensionality. We introduce ScoreCLR, a contrastive representation learning method that offers a novel way to estimate this repulsion term. We consider the framework of representation learning by maximizing the mutual information between two latent views of a sample. By deriving the gradient of the Shannon entropy with respect to the encoder parameters, we obtain an equivalent objective that involves the score of the latent representation. We then use denoising score matching techniques to estimate this score with a lightweight network that is trained jointly with the encoder. We further propose regularization techniques using different noise levels and a Gaussian prior on the latent representation. ScoreCLR replaces the batch of explicit negative samples in contrastive methods with a learned repulsion whose cost is linear in the batch size rather than quadratic. On CIFAR-10 and ImageNet-100 it does so with minimal loss of linear-probe accuracy compared to SimCLR under a shared recipe, while producing projector outputs of higher effective rank, closer to the Gaussian prior.

\end{abstract}

\section{Introduction}

The goal of representation learning is to learn features from data that make downstream tasks easier to solve \citep{goodfellow2016deep}. It is now widely used across several machine learning domains, including generative modeling \citep{rombach2022high}, reinforcement learning \citep{hafner2024mastering}, and causality \citep{scholkopf2021causal}. Representation learning methods can be broadly divided into two categories: reconstructive approaches \citep{vincent2010stacked, kingma2013auto, he2021masked}, which train by reconstructing samples from their latent encodings, and non-reconstructive ones \citep{chen2020simple, caron2021emerging, assran2023ijepa}, which optimize a training criterion in the latent space.

For the latter, two main approaches are typically considered. Contrastive methods \citep{oord2018representation, chen2020simple} bring representations of different views of the same sample closer while separating those of different samples. Estimating the repulsion force is generally challenging: it relies on in-batch negatives whose cost grows quadratically with the batch size, suffers from high variance and bias \citep{poole2019variational}, and requires careful hyperparameter tuning \citep{chen2021empirical}. This has led to the development of non-contrastive methods that focus solely on aligning latent representations and prevent collapse through architectural design and training asymmetry \citep{grill2020bootstrap, simsiam2021chen, caron2021emerging, assran2023ijepa} or explicit regularization in the training objective \citep{bardes2022vicreg, balestriero2025lejepa}. Although these techniques do prevent collapse \citep{ponce2026dual}, it is unclear which structure they impose on the latent space, as well as the efficiency of the estimator they rely on (see Section~\ref{sec:regularization-comparison}).

We introduce ScoreCLR, a novel method for contrastive representation learning that estimates the repulsion term using denoising score matching techniques \citep{vincent2011connection, song2019generative}. Our training criterion derives from maximizing the mutual information between two views of a sample \citep{tschannen2020mutual}. By deriving the gradient of the Shannon entropy with respect to the encoder parameters, we construct an equivalent loss, sharing the same gradient, that involves the score of the latent representation $\nabla_z \log p_Z(z)$. We then learn this score across different noise levels using a lightweight score network trained jointly with the encoder following the methodology of \citet{yin2024improved}. We call this score-based estimate of the entropy term of the mutual information ScoreReg, by analogy with VICReg \citep{bardes2022vicreg} and SIGReg \citep{balestriero2025lejepa}.

\begin{figure}[ht]
\centering
\includegraphics[width=\textwidth]{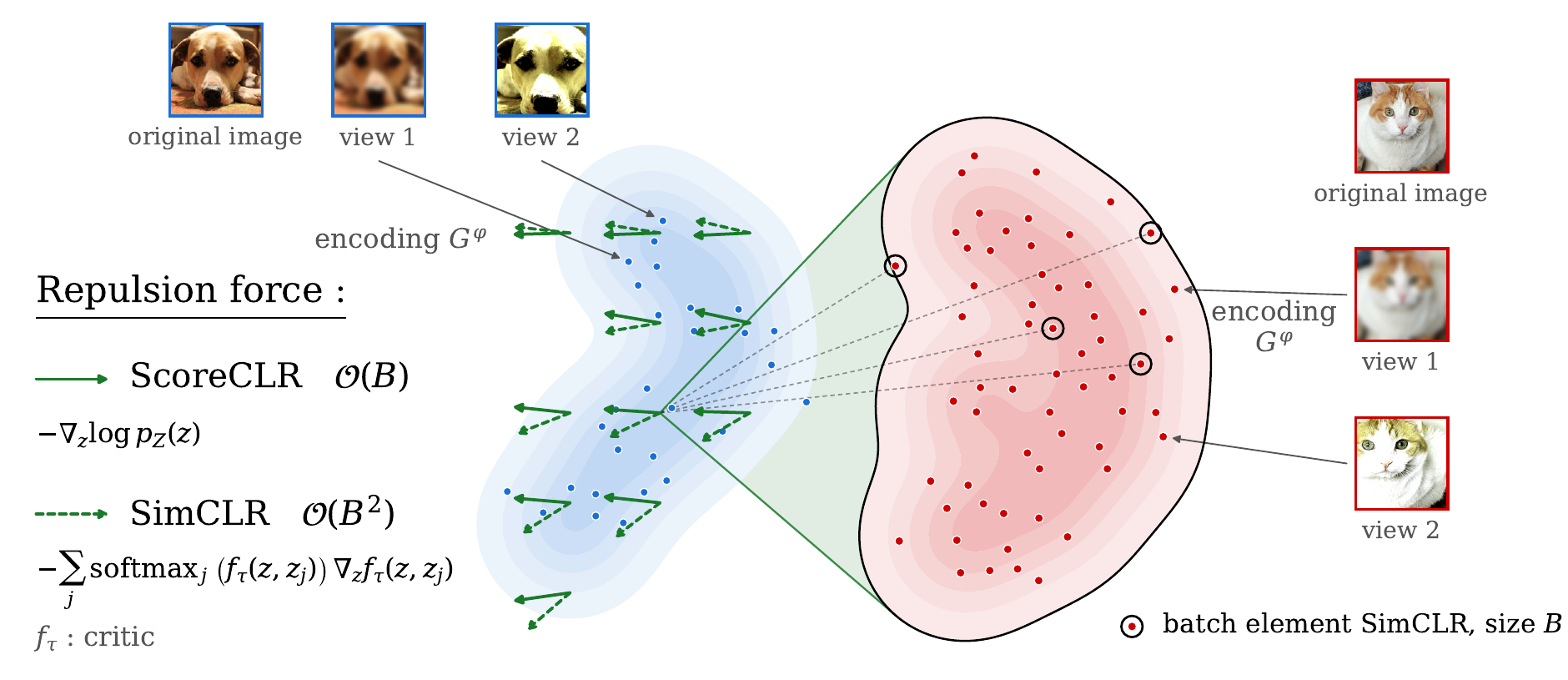}
\caption{Repulsion force acting on a latent code $z$ (negative gradient of the loss). SimCLR (dashed) estimates it against the $B$ codes of the batch, with a cost $\mathcal{O}(B^2)$ per batch, whereas ScoreCLR (solid) uses the score $-\nabla_z \log p_Z(z)$ of the latent distribution, learned by a small network, with a cost $\mathcal{O}(B)$.}
\label{fig:repulsion}
\end{figure}

Unlike previous contrastive methods \cite{chen2020simple,he2020momentum} whose loss estimation is quadratic in the batch size, ours is linear: the repulsion is amortized in a score network instead of being computed against a batch (Figure~\ref{fig:repulsion}). Empirically, ScoreCLR reaches competitive linear-probe accuracy on CIFAR-10 and ImageNet-100, while yielding projector outputs of higher effective rank than SimCLR and LeJEPA, which span only a fraction of the dimensions. The code is available at \url{https://github.com/chris-mrn/ScoreCLR_code}.

\section{Representation learning via mutual information maximization}

Let $X$ denote the random variable representing the data. We consider learning representations by learning to distinguish the joint distribution of two different latent views of $X$ from the product of their marginals. More formally, let $T$ denote a stochastic transformation of the data (e.g., cropping, masking, blurring, flipping, etc.) and $T_1$, $T_2$ two i.i.d.\ copies of $T$. We will consider the latent representation to be $Z = G(T(X))$, where $G$ is an encoding function and denote $Z_1 =G(T_1(X))$ and $Z_2 = G(T_2(X))$ two latent views of $X$. The objective is to maximize
\[
\text{D}(p_{(Z_1, Z_2)}||p_{Z_1}  p_{ Z_2}),
\]
where $D$ is a divergence criterion. Several divergences have been explored in the literature for learning representations such as Wasserstein \citep{ozair2019wasserstein}, Maximum Mean Discrepancy (MMD) \citep{gretton2005measuring, li2021self}, and Jensen--Shannon \citep{hjelm2019learning}. In our case we will consider the Kullback--Leibler ($\text{KL}$) divergence that inherits from desirable statistical properties \citep{cover2006elements}. We are then in the InfoMax principle \citep{linsker1988self} framework and our objective is to maximize
\[
I(Z_1, Z_2) = \text{KL}(p_{(Z_1, Z_2)}||p_{Z_1}  p_{ Z_2}) = H(Z_1) - H(Z_1 \mid Z_2).
\]
 Using this decomposition, we can interpret the objective as simultaneously encouraging the representation to have high entropy and making it predictable from the other representation view, through minimizing the conditional entropy $H(Z_1 \mid Z_2)$. We will parametrize the encoding function with a deep neural network $G^\varphi$, denote $Z_1^\varphi = G^\varphi(T_1(X))$, $Z_2^\varphi = G^\varphi(T_2(X))$, and aim at minimizing
\[
\mathcal{L}_{\text{MI}}(\varphi) =   H(Z_1^\varphi \mid Z_2^\varphi)- H(Z_1^\varphi).
\]

We will now present different ways to estimate the mutual information loss that can be found in the literature.

\subsection{Variational bound and kernel density estimator}
A common way to construct an estimator is to consider a lower bound using a dual representation of the KL. A standard choice is the Donsker--Varadhan representation, where the supremum is taken over measurable functions $f$ such that both expectations below are finite, which gives
\[
\text{KL}(p_{(Z_1, Z_2)}\,\|\,p_{Z_1}  p_{Z_2})  = \sup_{f}
\left\{
\mathbb{E}\left[f\left(Z_1, Z_2\right)\right]
-
\log \mathbb{E}\left[e^{f\left(Z_1, \tilde{Z}_2\right)}\right]
\right\},
\]
where $\tilde{Z}_2 = G^\varphi(T_2(\tilde{X}))$ with $\tilde{X}$ independent of $X$, so that $(Z_1, \tilde{Z}_2) \sim p_{Z_1} p_{Z_2}$.
Any $f$ thus gives a lower bound on the mutual information.
Let $f_{\tau}(x,y) = \frac{1}{\tau} \frac{x^{\top} y}{\|x\| \|y\|}$. We consider the critic
\[
f(z_1, z_2) = f_{\tau}(z_1, z_2) - \log \mathbb{E}_{\tilde{Z}_2}\Big[e^{f_{\tau}(z_1, \tilde{Z}_2)}\Big].
\]
Since $\tilde{Z}_2$ is independent of $Z_1$, $\mathbb{E}_{Z_1, \tilde{Z}_2}\big[e^{f(Z_1, \tilde{Z}_2)}\big] = 1$, and the logarithm term of the Donsker--Varadhan bound vanishes.
Using Monte Carlo estimators of the expectations with i.i.d.\ samples $(Z_1^i, Z_2^i)_{i \in \{1, \cdots, B\}}$ of $(Z_1, Z_2)$ and $(\tilde{Z}_2^j)_{j \in \{1, \cdots, B\}}$ of $\tilde{Z}_2$, we obtain the loss
\[
\mathcal{L}
=
-\frac{1}{B}
\underbrace{\sum_{i=1}^{B} f_{\tau}(Z_1^{i},Z_2^{i})}_{\text{attraction}}+
 \underbrace{\frac{1}{B}\sum_{i=1}^{B} \log \left(\frac{1}{B}
\sum_{j=1}^{B}
\exp\!\left(f_{\tau}(Z_1^{i},\tilde{Z}_2^{j})\right) \right)}_{\text{repulsion}}
.
\]
This loss is closely related to the InfoNCE loss of \citet{oord2018representation}, the difference being that the positive pair $(Z_1^i, Z_2^i)$ does not appear in the denominator, as in decoupled contrastive learning \citep{yeh2022decoupled}.

Up to an additive constant, the repulsion term is minus a plug-in kernel density estimator of the entropy of the normalized latent on the sphere, with the kernel $k_{\tau}(x, y) = \exp(f_{\tau}(x,y))$.
Minimizing it encourages the latent representation to have high entropy, which favors uniformity on the sphere \citep{wang2020understanding}.

The SimCLR loss \citep{chen2020simple} is a close variant, which symmetrizes the loss with respect to the two views, uses the other samples of the batch from both views as negatives, and includes the positive pair in the denominator.
Other works modify the samples used in this estimator. NNCLR \citep{dwibedi2021little} replaces $Z_1^i$ by its nearest neighbour in a queue of embeddings stored from previous batches, and MoCo \citep{he2020momentum} draws negatives from a dynamic memory queue, which decouples the number of negatives from the batch size and enables the use of smaller batch sizes.

These estimators all suffer from the curse of dimensionality under standard regularity conditions on the density of interest \citep{scott1991feasibility, lecun2022path}. The relevant dimension is in general the intrinsic dimension of the data rather than the ambient one, but it has been recently suggested that this intrinsic dimension is still too large for standard statistical estimators on real-world data \citep{pope2021intrinsic}. Contrary to classic kernel density estimators, which only leverage a weak prior on the distribution of interest, the ScoreCLR repulsion estimator includes a neural prior that may benefit from a better sample complexity \citep{barron1993universal}.

\subsection{Neural methods}

Several approaches have been proposed to estimate or maximize mutual information in representation learning relying on neural critics. For example, MINE \citep{belghazi2018mine} used the dual representation of the $\text{KL}$ with a critic learned by a neural network $f_\psi$, which was subsequently used for unsupervised representation learning in Deep InfoMax \citep{hjelm2019learning}. The mutual-information-based loss can be rewritten as
\[
\mathcal{L}(\varphi) = \sup_{\psi}
\left\{
\mathbb{E}\left[
f_\psi\left(Z_1^\varphi, Z_2^\varphi\right)
\right]
-
\log \mathbb{E}_{}\left[
e^{f_\psi\left(Z_1^\varphi, \tilde{Z}_2^\varphi\right)}
\right]
\right\},
\]
where $\tilde{Z}_2^\varphi = G^\varphi(T_2(\tilde{X}))$ with $\tilde{X}$ independent from $X$.

Training then proceeds by alternating maximization over the critic parameters \(\psi\) and maximization of the resulting bound over the encoder parameters \(\varphi\). An estimator of the mutual information during training is given by
\[
\widehat{I}_{\text{DV}}(\varphi, \psi) = \frac{1}{B} \sum_{i=1}^B f_\psi\left(Z_{1,i}^\varphi, Z_{2,i}^\varphi\right) - \log \left( \frac{1}{B} \sum_{i=1}^B e^{f_\psi\left(Z_{1,i}^\varphi, \tilde{Z}_{2,i}^\varphi\right)} \right).
\]

In practice, these variational estimators can suffer from optimization instability and can lead to collapsed representations \citep{chae2023localminimax}. Moreover, these neural estimators do not necessarily preserve desirable properties of the underlying KL divergence, such as invariance to reparameterization \citep{song2020understanding}.

\section{ScoreCLR}
More recently, diffusion-based approaches have been proposed for mutual information estimation \citep{franzese2024minde}. These methods achieve improved performance and pass the invariance to reparameterization test \citep{kholkin2025infobridge}.
To the best of our knowledge, these diffusion-based approaches
to mutual information estimation have not yet been explored in a
self-supervised representation learning framework.

\subsection{Equivalent loss}
Since we optimize our loss with gradient descent, replacing it by another loss with the same gradient does not change the training dynamic. The following proposition gives a way to make such a replacement for the entropy term.
\begin{proposition}
\label{prop:entropy-grad}
For any random variable $Y$, denoting $Z^\varphi = F^\varphi(Y)$, under Assumption~\ref{ass:entropy-grad} (Appendix~\ref{app:proofs}) we have
\[
\nabla_\varphi H(Z^\varphi) = \frac{1}{2}\nabla_{\varphi}\mathbb{E}[\| Z^\varphi - \text{sg} ( Z^\varphi +\nabla_z \log p_{Z^{\varphi}}(Z^{\varphi}))\|^2],
\]
where $\text{sg}$ denotes the stop grad operator.
\end{proposition}

For the conditional entropy, we will consider the Gaussian conditional model $Z_1 \mid Z_2 \sim \mathcal{N}(Z_2, \sigma^2 I)$ which will give a lower bound on the mutual information,
\[
\mathcal{L}_{\text{MI}}(\varphi)  = \frac{1}{2\sigma^2}\mathbb{E}[\|G^\varphi(T_1(X))-G^\varphi(T_2(X))\|^2] - H(Z^\varphi),
\]

that gives the same training dynamic as
\[
\mathcal{L}_{\text{MI-score}}(\varphi) = \frac{1}{2\sigma^2}\mathbb{E}[\|G^\varphi(T_1(X))-G^\varphi(T_2(X))\|^2] + \underbrace{\frac{1}{2}\mathbb{E}[\| Z^\varphi - \text{sg} ( Z^\varphi -\nabla_z \log p_{Z^{\varphi}}(Z^{\varphi}))\|^2]}_{\text{ScoreReg}}.
\]

We call its second term ScoreReg. This loss can be seen as enforcing invariance of the latent representation to the given transformation while encouraging it to remain of high entropy.

\subsection{Regularization}

\paragraph{Gaussian prior}
Without any extra regularization the problem is only constrained by the architecture of the network and the entropy of the latent could grow arbitrarily high. To alleviate this issue we consider a Gaussian prior on our latent variable by adding an $L^2$ penalty term to the loss. Noticing that $-H(Z) + \frac{1}{2}\mathbb{E}[{\|Z\|^2}] = C + \text{KL}(p_Z\,\|\,\mathcal{N}(0,I))$, we recover the same regularization term as in \citet{balestriero2025lejepa}, the maximum-entropy prior with mean \(0\) and variance \(1\). In practice the penalty is applied to the noised latents at which the score is queried (Algorithm~\ref{alg:scoreclr}).

\paragraph{Noise regularization}
To estimate the score of the latent we use denoising score matching \citep{vincent2011connection, song2019generative}: we approximate the score of the latent distribution convolved with Gaussian noise of standard deviation $t$. We denote $Z_t^\varphi = Z^\varphi + t\varepsilon$, where $\varepsilon \sim \mathcal{N}(0, I)$, and train a lightweight noise-prediction network $\varepsilon^\theta$ with the denoising loss
\[
\mathcal{L}_{\text{DSM}}(\theta) = \mathbb{E}\big[\|\varepsilon^\theta(Z_t^\varphi, t) - \varepsilon\|^2\big], \qquad t \sim p(t),
\]
where $p(t)$ is the noise level distribution, specified in Section~\ref{sec:setup}. The score is recovered by Tweedie's formula, $s_t^\theta(z) = -\varepsilon^\theta(z, t)/t$. Further details can be found in Appendix~\ref{app:details}.

For simplicity, we replace the $1/(2\sigma^2)$ factor by a tunable weight $\lambda$. The regularized loss is
\[
\mathcal{L}_{\text{MI}}^{\text{reg}}(\varphi) = \lambda\mathbb{E}\big[\|Z_1^\varphi-Z_2^\varphi\|^2\big] + \mathbb{E}_{t\sim p(t)}\Big[\text{KL}\big(p_{Z_t^\varphi} \,\|\, \mathcal{N}(0, I)\big)\Big].
\]

Using Proposition~\ref{prop:entropy-grad}, $\mathcal{L}_{\text{MI}}^{\text{reg}}$ will give the same training dynamic as
\[
\mathcal{L}_{\text{MI-score}}^{\text{reg}}(\varphi) = \lambda\mathbb{E}\big[\|Z_1^\varphi-Z_2^\varphi\|^2\big] + \frac{1}{2}\,\mathbb{E}_{t\sim p(t)}\,\mathbb{E}\Big[\big\|Z^\varphi - \text{sg}\big(Z^\varphi - \nabla_z \log p_{Z_t^\varphi}(Z_t^\varphi) - Z_t^\varphi\big)\big\|^2\Big].
\]

To have a low variance estimate of the repulsive term, we may average it over $K$ noise levels, using independent Gaussian noise for each level.

\subsection{Training}
We want, for every update of $\varphi$, a statistical estimator of the score $s_t^{\theta^{*}(\varphi)}$ that minimizes well $\mathcal{L}_{\text{DSM}}(\theta)$. To do so, we follow the methodology of distribution matching distillation \citep{yin2024improved}, which alternates training between updates of $\varphi$ and $\theta$ with a 1-to-$k$ proportion rule that has proven to be efficient in large-scale training. The score network is unconditional: it takes only the noised code and the noise level.

The final loss can be made of three components, an invariance term $\mathcal{L}_{\text{inv}}$, a repulsion term $\mathcal{L}_{\text{rep}}$ that estimates ScoreReg, and a regularization term $\mathcal{L}_{\text{reg}}$, the last two forming the prior term $\mathcal{L}_{\text{prior}}$. On a batch, their expressions are
\begin{equation}
\label{eq:scoreclr_loss}
\begin{aligned}
\mathcal{L}_{\text{inv}} &= \frac{1}{Bd}\sum_{i=1}^{B}\|Z_1^{i}-Z_2^{i}\|^2,\\
\mathcal{L}_{\text{rep}} &= \frac{1}{2Bd}\sum_{i=1}^{B}\sum_{v \in \{1,2\}}\frac{1}{2}\big\|Z_v^{i}-\text{sg}\big(Z_v^i-\bar{s}_v^{i}\big)\big\|^2,\\
\mathcal{L}_{\text{reg}} &= \frac{1}{2Bd}\sum_{i=1}^{B}\sum_{v \in \{1,2\}}\frac{1}{K}\sum_{\ell=1}^{K}\frac{1}{2}\|Z_{v,\ell}^{i}\|^2,\\
\mathcal{L}_{\text{ScoreCLR}} &= \,\lambda\mathcal{L}_{\text{inv}} + \, \mathcal{L}_{\text{prior}}, \quad
\mathcal{L}_{\text{prior}} = \mathcal{L}_{\text{rep}} + \mathcal{L}_{\text{reg}},
\end{aligned}
\end{equation}
where $B$ is the batch size, $d$ is the dimension of the codes, $Z_v^{i} = G^\varphi(T_v(X^{i}))$, $Z_{v,\ell}^{i} = Z_v^{i} + t_{v,\ell}^{i}\varepsilon_{v,\ell}^{i}$ with $t_{v,\ell}^{i} \sim p(t)$ and $\varepsilon_{v,\ell}^{i} \sim \mathcal{N}(0,I)$ drawn independently for each sample, view and query, and $\bar{s}_v^{i} = \frac{1}{K}\sum_{\ell=1}^{K} s_{t_{v,\ell}^{i}}^\theta(Z_{v,\ell}^{i})$.
The training procedure is summarized in Algorithm~\ref{alg:scoreclr}.
Connections to the SimCLR loss and to drifting generative models \citep{deng2026drifting} are given in Appendices~\ref{app:SimCLRvsScoreCLR} and~\ref{app:drifting}, respectively.

\begin{algorithm}[H]
\caption{ScoreCLR training}
\label{alg:scoreclr}
\begin{algorithmic}[1]
\Require Encoder $G^\varphi$, noise-prediction network $\varepsilon^\theta$, augmentations $T_1,T_2$, noise distribution $p(t)$, score-network steps $S$, noise queries $K$, weight $\lambda$
\For{each batch $(X^i)_{i=1}^{B}$}
\State $Z_v^i = G^\varphi(T_v(X^i))$ for $v \in \{1,2\}$
\For{$j = 1, \dots, S$} \Comment{score-network updates}
\State Sample new $t_v^{i,(j)} \sim p(t)$ and $\varepsilon_v^{i,(j)} \sim \mathcal{N}(0,I)$ for every $i$ and $v$
\State $\widetilde{Z}_v^{i,(j)} = \text{sg}(Z_v^i) + t_v^{i,(j)}\,\varepsilon_v^{i,(j)}$
\State $\theta \gets$ 1 update of $\displaystyle\frac{1}{2B}\sum_{i=1}^{B}\sum_{v\in\{1,2\}}\big\|\varepsilon^\theta\big(\widetilde{Z}_v^{i,(j)}, t_v^{i,(j)}\big)-\varepsilon_v^{i,(j)}\big\|^2$
\EndFor
\State Sample new $t_{v,\ell}^i \sim p(t)$, $\varepsilon_{v,\ell}^i \sim \mathcal{N}(0,I)$, $\ell = 1,\dots,K$, and compute $Z_{v,\ell}^i$ and $\bar{s}_v^i$ with the updated $\theta$
\State Compute $\mathcal{L}_{\text{ScoreCLR}}$ with Eq.~(\ref{eq:scoreclr_loss})
\State $\varphi \gets$ 1 update of $\mathcal{L}_{\text{ScoreCLR}}$
\EndFor
\end{algorithmic}
\end{algorithm}

\subsection{Comparison with other regularization methods}
\label{sec:comparison}
\label{sec:regularization-comparison}

Several explicit regularization terms have been proposed to prevent collapse in non-contrastive self-supervised learning. VICReg \citep{bardes2022vicreg} regularizes the variance and the covariance of the latent and thus implicitly assumes that the latent is Gaussian. SIGReg \citep{balestriero2025lejepa} pushes the latent toward an isotropic Gaussian with an Epps--Pulley characteristic-function test on random one-dimensional projections, relying on the Cram\'er--Wold theorem. As for every sliced divergence, the sample complexity of this test does not depend on the dimension \citep{nadjahi2020statistical}, but its statistical power does. Under concentration assumptions on the latent, most one-dimensional projections of a high-dimensional distribution are close to Gaussian \citep{diaconis1984asymptotics}, and the fraction of directions in which a projection deviates from Gaussian by more than $\epsilon$ decays as $\exp(-c\,d\,\epsilon^2)$ \citep{meckes2009quantitative}, so the number of random slices needed to detect such a deviation grows exponentially with the dimension, which recovers the curse of dimensionality in the statistical power of the test. A latent supported on a low-dimensional subspace can thus pass the test, which is the geometric collapse we observe for LeJEPA in Section~\ref{sec:geometry}.

ScoreCLR instead derives its repulsion from the contrastive criterion itself, as the gradient of the entropy term of the mutual information estimated with the full score of the latent, and does not suffer from this geometric collapse (Section~\ref{sec:geometry}).

\paragraph{Concurrent work.} We recently became aware of the concurrent and independent work of \citet{kappiyath2026ebreg}, which also trains a latent score network jointly with the encoder and uses the same identity between the gradient of the entropy and the score (Proposition~\ref{prop:entropy-grad}) to prevent collapse. They report strong results on ImageNet-1K with ResNet and ViT encoders, and also use the learned score at test time to adapt the representations of corrupted images. The two methods differ in how the latent is constrained, how the score is learned and how the entropy term is written. \citet{kappiyath2026ebreg} normalize the embeddings, learn the score at a single noise level with 16 score-network updates per encoder step, and use the linear surrogate $\mathrm{sg}(s)^\top \tilde z$. ScoreCLR constrains the latent with a soft Gaussian prior, which gives the KL divergence to $\mathcal{N}(0, I)$, learns the score over a distribution of noise levels with one to five score-network updates per step, and writes the entropy term together with the prior as a regression towards a target moved along the drift field $-\nabla_z \log p - z$, the velocity of the Wasserstein gradient flow of $\mathrm{KL}(\cdot\,\|\,\mathcal{N}(0,I))$ (Appendix~\ref{app:drifting}); both forms give the same encoder gradient. We also study the geometry of the codes and the cost of the objective as the batch size grows (Sections~\ref{sec:geometry} and~\ref{sec:cost}).

\section{Experiments}

We compare ScoreCLR with SimCLR, and on CIFAR-10 with LeJEPA \citep{balestriero2025lejepa}, BYOL \citep{grill2020bootstrap} and Barlow Twins \citep{zbontar2021barlow}, on linear evaluation, representation geometry, out-of-domain transfer, batch size, numerical precision and training cost. The main results of SimCLR, LeJEPA, BYOL, Barlow Twins and ScoreCLR under AdamW are means $\pm$ standard deviations over three seeds (1, 2, 3); the SimCLR (LARS) rows and the batch-size study are single runs (seed 5). Unless stated otherwise, runs are in fp32; SimCLR (LARS) is trained in 16-bit mixed precision as in solo-learn \citep{dacosta2022sololearn}. ScoreCLR uses $\lambda=5$, one score-network step ($S=1$) and one noise query per code ($K=1$) on CIFAR-10, and $\lambda=2$, $S=5$ and $K=3$ on ImageNet-100 (Appendix~\ref{app:details}); the AdamW methods share the same augmentations and gradient clipping (norm 1.0).

\subsection{Setup}
\label{sec:setup}

\textbf{Data.} We pre-train on CIFAR-10
\citep{krizhevsky2009learning} and ImageNet-100 \citep{deng2009imagenet}.
Transfer targets are CIFAR-10, STL-10 \citep{coates2011stl10},
and DTD \citep{cimpoi2014dtd}.

\textbf{Models.} ResNet-18 with a two-layer MLP projector (2048 hidden units, 256 outputs, no BatchNorm: a BatchNorm layer improves the results but partly prevents collapse on its own \citep{richemond2020byol}, so we remove it). SimCLR, LeJEPA, BYOL, Barlow Twins and ScoreCLR share the encoder, the projector, the optimizer (AdamW), the augmentations and the gradient clipping (Table~\ref{tab:hparams}, Appendix~\ref{app:details}). We also report SimCLR with LARS under solo-learn's recipe \citep{dacosta2022sololearn}, with the shared 2048/256 head and its published 4096/512 head.

\textbf{Noise levels.} The noise levels of ScoreCLR follow $\ln t \sim \mathcal{N}(-1.2,\, 1.2^2)$ on CIFAR-10 and $\ln t \sim \mathcal{N}(-1.2,\, 1.0^2)$ on ImageNet-100.

\textbf{Evaluation.} (i) Offline linear probe and weighted $k$-NN ($k=20$) on frozen features; (ii) alignment and uniformity \citep{wang2020understanding} of the projector output; (iii) effective rank \citep{garrido2023rankme} of the backbone features and of the projector output, and an estimate of the Poincar\'e constant of the projector output; (iv) out-of-domain linear transfer following \citet{kornblith2019better}; (v) measured time and peak memory of the objective, of a training step and of a training epoch.

\subsection{Linear and $k$-NN evaluation}
\label{sec:linear}

\begin{table}[ht]
\centering
\caption{Linear-probe and weighted $k$-NN ($k=20$) accuracy (\%) on frozen features, ResNet-18. Top: shared optimizer and projector (AdamW, gradient clip 1.0, 2048/256 projector without BatchNorm); mean $\pm$ std over three seeds (top-5 averaged); BYOL and Barlow Twins use a projector and predictor without BatchNorm (Appendix~\ref{app:details}). Bottom: SimCLR trained with solo-learn's LARS recipe, with the shared 2048/256 projector and with the published 4096/512 projector. With the latter, our offline linear probe gives 77.02 on ImageNet-100, against 77.64 for the online probe of the published checkpoint.}
\label{tab:linear}
\small
\setlength{\tabcolsep}{3pt}
\resizebox{\textwidth}{!}{%
\begin{tabular}{lllcccccccc}
\toprule
& & & \multicolumn{4}{c}{CIFAR-10} & \multicolumn{4}{c}{ImageNet-100} \\
\cmidrule(lr){4-7} \cmidrule(lr){8-11}
& & & \multicolumn{2}{c}{Linear} & \multicolumn{2}{c}{$k$-NN} & \multicolumn{2}{c}{Linear} & \multicolumn{2}{c}{$k$-NN} \\
\cmidrule(lr){4-5} \cmidrule(lr){6-7} \cmidrule(lr){8-9} \cmidrule(lr){10-11}
Method & Optim. & Projector & Top-1 & Top-5 & Top-1 & Top-5 & Top-1 & Top-5 & Top-1 & Top-5 \\
\midrule
SimCLR          & AdamW & 2048/256 & $89.26{\pm}0.38$ & 99.70 & $88.73{\pm}0.18$ & 98.90 & $72.39{\pm}0.62$ & 92.18 & $66.41{\pm}0.44$ & 87.32 \\
LeJEPA          & AdamW & 2048/256 & $89.67{\pm}0.13$ & 99.65 & $88.40{\pm}0.13$ & 98.65 & -- & -- & -- & -- \\
BYOL            & AdamW & 2048/256 & $88.96{\pm}0.30$ & 99.70 & $88.44{\pm}0.22$ & 99.01 & -- & -- & -- & -- \\
Barlow Twins    & AdamW & 2048/256 & $88.05{\pm}0.15$ & 99.57 & $87.17{\pm}0.09$ & 98.57 & -- & -- & -- & -- \\
ScoreCLR (ours) & AdamW & 2048/256 & $88.87{\pm}0.44$ & 99.66 & $88.27{\pm}0.13$ & 98.85 & $71.23{\pm}0.06$ & 91.03 & $66.05{\pm}0.39$ & 86.09 \\
\midrule
SimCLR          & LARS  & 2048/256 & 90.79 & 99.74 & 89.99 & 99.02 & 75.40 & 93.30 & 69.98 & 88.64 \\
SimCLR          & LARS  & 4096/512 & -- & -- & -- & -- & 77.02 & 93.84 & 71.60 & 89.10 \\
\bottomrule
\end{tabular}%
}
\end{table}

\textbf{Discussion.} With the same optimizer and projector, ScoreCLR is competitive with the contrastive and joint-embedding baselines. On CIFAR-10 its linear accuracy is 0.4 points below SimCLR (within one standard deviation) and 0.8 points below LeJEPA, within 0.1 of BYOL and 0.8 above Barlow Twins, and its $k$-NN accuracy is within 0.5 points of SimCLR, LeJEPA and BYOL. On ImageNet-100 the linear probe is 1.2 points below SimCLR and the $k$-NN accuracy 0.4 points below, with a smaller seed spread (0.06 against 0.62). SimCLR with LARS is stronger than all AdamW methods; these rows come from solo-learn's own pipeline and are reference points, not a controlled comparison of objectives (Appendix~\ref{app:protocol}). The takeaway is that replacing the $\mathcal{O}(B^2)$ batch-based repulsion by a learned score costs 0.4 points of linear accuracy on CIFAR-10 and 1.2 on ImageNet-100: what the method offers lies in the geometry of $z$ and in the cost of the objective at large batch sizes (Sections~\ref{sec:geometry} and~\ref{sec:cost}), not in the probe accuracy.

\subsection{Representation Geometry}
\label{sec:geometry}

Table~\ref{tab:geometry} reports alignment and uniformity of the projector output and Table~\ref{tab:rank} the effective rank \citep{garrido2023rankme} of the backbone features $h$ and of the projector output $z$. In $z$, SimCLR spans about half of the 256 dimensions on ImageNet-100 and a third on CIFAR-10, the dimensional collapse of InfoNCE described by \citet{jing2022understanding}.  LeJEPA has the lowest rank in both $h$ (253/512) and $z$ (58/256) although it is trained with the SIGReg criterion (Section~\ref{sec:comparison}). ScoreCLR's $z$ has a much higher effective rank (229/256 on CIFAR-10 and 244/256 on ImageNet-100), and its uniformity equals SimCLR's on CIFAR-10 and is better on ImageNet-100 ($-3.93$ against $-3.86$). It pays for this with a worse alignment (0.51 against 0.37 for SimCLR on CIFAR-10, and 0.38 against 0.24 on ImageNet-100): the positive pairs are held less tightly together. The takeaway is that the score-based repulsion spreads the codes over almost all the available dimensions, which neither InfoNCE nor the SIGReg regularizer does. The ranks of the backbone features $h$ are similar for ScoreCLR and SimCLR on CIFAR-10 (332 and 336/512) and slightly higher for ScoreCLR on ImageNet-100 (346 against 329/512).

Table~\ref{tab:rank} also reports an estimate $\widehat{C}_P$ of the Poincar\'e constant of the codes. The Poincar\'e constant $C_P$ of a distribution $p$ is the smallest constant such that $\mathrm{Var}_p(f)\le C_P\,\mathbb{E}_p\|\nabla f\|^2$ for every smooth function $f$ \citep{bakry2014markov}. It equals $1$ for the standard Gaussian, and it is large when the distribution has separated modes or strongly anisotropic directions: it is the inverse of the spectral gap of the Langevin diffusion that targets $p$, so it controls how fast that diffusion mixes \citep{bakry2014markov} and the statistical efficiency of learning the score of $p$ by score matching \citep{koehler2023statistical}. It is therefore a natural measure of how close the geometry of $z$ is to the Gaussian prior of ScoreCLR. We estimate it with the random-feature estimator of \citet{pillaudvivien2020statistical} (Appendix~\ref{app:poincare}). With this estimator, on whitened codes, the value for ScoreCLR is the smallest of the three methods on CIFAR-10 (it was not computed on ImageNet-100).

%A higher rank is not by itself a proxy for accuracy: on the backbone features used by the probe, a lower intrinsic dimension correlates with higher linear-probe accuracy across SSL methods \citep{mordacq2026idest}; our claim only concerns $z$.

\begin{table}[ht]
\centering
\caption{Alignment ($\downarrow$) and uniformity ($\downarrow$)
\citep{wang2020understanding} of the projector output, ResNet-18 (AdamW rows: mean $\pm$ std over three seeds; LARS rows: single run). ``--'': not measured.}
\label{tab:geometry}
\small
\begin{tabular}{llcc cc}
\toprule
& & \multicolumn{2}{c}{CIFAR-10} & \multicolumn{2}{c}{ImageNet-100} \\
\cmidrule(lr){3-4} \cmidrule(lr){5-6}
Method & Optim. & Align & Unif & Align & Unif \\
\midrule
SimCLR          & LARS  & 0.349 & $-3.818$ & 0.240 & $-3.856$ \\
SimCLR          & AdamW & $0.367{\pm}0.004$ & $-3.804{\pm}0.002$ & $0.235{\pm}0.001$ & $-3.855{\pm}0.001$ \\
LeJEPA          & AdamW & $0.316{\pm}0.003$ & $-3.540{\pm}0.003$ & -- & -- \\
ScoreCLR (ours) & AdamW & $0.514{\pm}0.006$ & $-3.802{\pm}0.001$ & $0.378{\pm}0.005$ & $-3.931{\pm}0.000$ \\
\bottomrule
\end{tabular}
\end{table}

\begin{table}[ht]
\centering
\caption{Effective rank (RankMe, \citealp{garrido2023rankme}) of the backbone features $h$ and of the projector output $z$ on raw (un-normalised) validation features, out of the space dimension, and an estimate $\widehat{C}_P$ of the Poincar\'e constant of the whitened $z$ (1 for a Gaussian; lower is closer to Gaussian), estimated with the random-feature estimator of \citet{pillaudvivien2020statistical} (Appendix~\ref{app:poincare}). Both datasets use a ResNet-18. AdamW rows: mean $\pm$ std over three seeds; LARS rows: single run. The low rank of $z$ under InfoNCE is the dimensional collapse analysed by \citet{jing2022understanding}.}
\label{tab:rank}
\small
\begin{tabular}{lllccc}
\toprule
Pre-training & Method & Optim. & RankMe $h$ (/512) & RankMe $z$ (/256) & $\widehat{C}_P$ of $z$ \\
\midrule
CIFAR-10     & SimCLR          & LARS  & 372 & 88 & -- \\
ResNet-18    & SimCLR          & AdamW & $336{\pm}2$ & $96{\pm}1$ & $25.0{\pm}1.4$ \\
             & LeJEPA          & AdamW & $253{\pm}1$ & $58{\pm}0$ & $13.5{\pm}1.5$ \\
             & BYOL            & AdamW & $292{\pm}8$ & $98{\pm}5$ & -- \\
             & Barlow Twins    & AdamW & $243{\pm}3$ & $73{\pm}0$ & -- \\
             & ScoreCLR (ours) & AdamW & $332{\pm}2$ & $229{\pm}0$ & $7.8{\pm}0.8$ \\
\midrule
ImageNet-100 & SimCLR          & LARS  & 368 & 118 & -- \\
ResNet-18    & SimCLR          & AdamW & $329{\pm}1$ & $143{\pm}0$ & -- \\
             & ScoreCLR (ours) & AdamW & $346{\pm}1$ & $244{\pm}0$ & -- \\
\bottomrule
\end{tabular}
\end{table}

\subsection{Transfer}

Table~\ref{tab:transfer_linear} probes frozen features on unseen datasets, STL-10 and DTD. ScoreCLR transfers slightly worse than SimCLR and is close to LeJEPA. After CIFAR-10 pre-training, it is 0.9 points below SimCLR on STL-10 and 2.5 below on DTD, and close to LeJEPA (75.2 against 76.0 on STL-10, 28.0 against 27.0 on DTD). After ImageNet-100 pre-training the gap to SimCLR is larger (1.5 points on STL-10, 3.1 on DTD, 4.4 on CIFAR-10), in line with the lower linear accuracy of Table~\ref{tab:linear}. The takeaway is that the higher effective rank of $z$ does not by itself make the backbone features transfer better: transfer follows the in-domain accuracy of the representation.

\begin{table}[ht]
\centering
\caption{Out-of-domain linear transfer, top-1 (\%): logistic regression on frozen L2-normalized features at the pretraining resolution (Appendix~\ref{app:transfer}). ``$\cdot$'' is the pretraining dataset (Table~\ref{tab:linear}). All rows: mean $\pm$ std over three seeds.}
\label{tab:transfer_linear}
\small
\setlength{\tabcolsep}{5pt}
\begin{tabular}{llccc}
\toprule
Method & Optim. & CIFAR-10 & STL-10 & DTD \\
\midrule
\multicolumn{5}{l}{\emph{Pretrained on CIFAR-10 (ResNet-18, 32\,px)}} \\
SimCLR          & AdamW & $\cdot$ & $76.12{\pm}0.81$ & $30.46{\pm}0.49$ \\
LeJEPA          & AdamW & $\cdot$ & $76.01{\pm}0.54$ & $27.02{\pm}1.99$ \\
BYOL            & AdamW & $\cdot$ & $77.07{\pm}0.54$ & $31.35{\pm}0.58$ \\
Barlow Twins    & AdamW & $\cdot$ & $74.94{\pm}0.42$ & $28.94{\pm}0.67$ \\
ScoreCLR (ours) & AdamW & $\cdot$ & $75.21{\pm}0.13$ & $28.01{\pm}0.53$ \\
\midrule
\multicolumn{5}{l}{\emph{Pretrained on ImageNet-100 (ResNet-18, 224\,px)}} \\
SimCLR          & AdamW & $75.64{\pm}0.30$ & $84.35{\pm}0.42$ & $59.38{\pm}1.18$ \\
ScoreCLR (ours) & AdamW & $71.23{\pm}0.36$ & $82.87{\pm}0.40$ & $56.28{\pm}0.69$ \\
\bottomrule
\end{tabular}
\end{table}

\subsection{Batch size}
\label{sec:batch}

We train SimCLR and ScoreCLR at larger batches (Table~\ref{tab:batch}). The encoder learning rate follows the square-root rule ($\mathrm{lr}\propto\sqrt{B/128}$) for both methods; for ScoreCLR the score-network learning rate is kept fixed. ScoreCLR stays within 0.9 points of SimCLR up to $B=2048$ and is 1.4 points below at $B=4096$. With a larger batch there are fewer iterations per epoch, and so fewer score-network updates. In our runs, the number $S$ of score-network updates per step had to grow with the batch size to keep the accuracy (Appendix~\ref{app:ablations}). In these experiments, ScoreCLR does not reduce the batch size needed for a given accuracy; at large batches, its main benefit is the lower cost of the objective (Section~\ref{sec:cost}).

\begin{table}[ht]
\centering
\caption{Linear accuracy (\%), $k$-NN accuracy (\%) and effective rank of $z$ against the batch size, CIFAR-10, 1000 epochs, seed 5. SimCLR: AdamW, gradient clipping, encoder learning rate scaled by $\sqrt{B/128}$. ScoreCLR: $K{=}1$, gradient clipping, same encoder scaling, unscaled score-network learning rate, $S{=}1$ at $B{=}128$ (main recipe), $S{=}5$ at $B{=}512$ and $S{=}10$ at $B{=}2048$ and $4096$.}
\label{tab:batch}
\small
\begin{tabular}{lcccccc}
\toprule
& \multicolumn{3}{c}{ScoreCLR} & \multicolumn{3}{c}{SimCLR} \\
\cmidrule(lr){2-4} \cmidrule(lr){5-7}
$B$ & Linear & $k$-NN & RankMe $z$ & Linear & $k$-NN & RankMe $z$ \\
\midrule
128  & 88.80 & 88.51 & 229 & 89.45 & 88.56 & 96 \\
512  & 88.51 & 88.39 & 231 & 89.32 & 89.07 & 108 \\
2048 & 88.29 & 87.76 & 232 & 88.63 & 88.92 & 107 \\
4096 & 87.50 & 86.88 & 230 & 88.88 & 88.78 & 107 \\
\bottomrule
\end{tabular}
\end{table}

\subsection{Numerical precision}
\label{sec:precision}

SimCLR (LARS) is trained in 16-bit mixed precision (solo-learn), while the other runs use fp32. ScoreCLR works equally well in fp16: over three seeds, fp16 training reaches $88.82{\pm}0.04$\% linear accuracy against $88.87{\pm}0.44$\%  in fp32, with the same effective rank of $z$ (Table~\ref{tab:precision}). The comparison with the mixed-precision SimCLR (LARS) baseline is therefore not biased by the precision.

\begin{table}[ht]
\centering
\caption{ScoreCLR on CIFAR-10 in fp32 and fp16 mixed precision (encoder and objective under autocast, score-network updates in fp32); mean $\pm$ std over seeds 1--3.}
\label{tab:precision}
\small
\begin{tabular}{lccc}
\toprule
Precision & Linear & $k$-NN & RankMe $z$ \\
\midrule
fp32 & $88.87{\pm}0.44$ & $88.27{\pm}0.13$ & $229{\pm}0$ \\
fp16 & $88.82{\pm}0.04$ & $88.39{\pm}0.05$ & $229{\pm}1$ \\
\bottomrule
\end{tabular}
\end{table}

\subsection{Computational Cost}
\label{sec:cost}

\begin{figure}[ht]
\centering
\includegraphics[width=\textwidth]{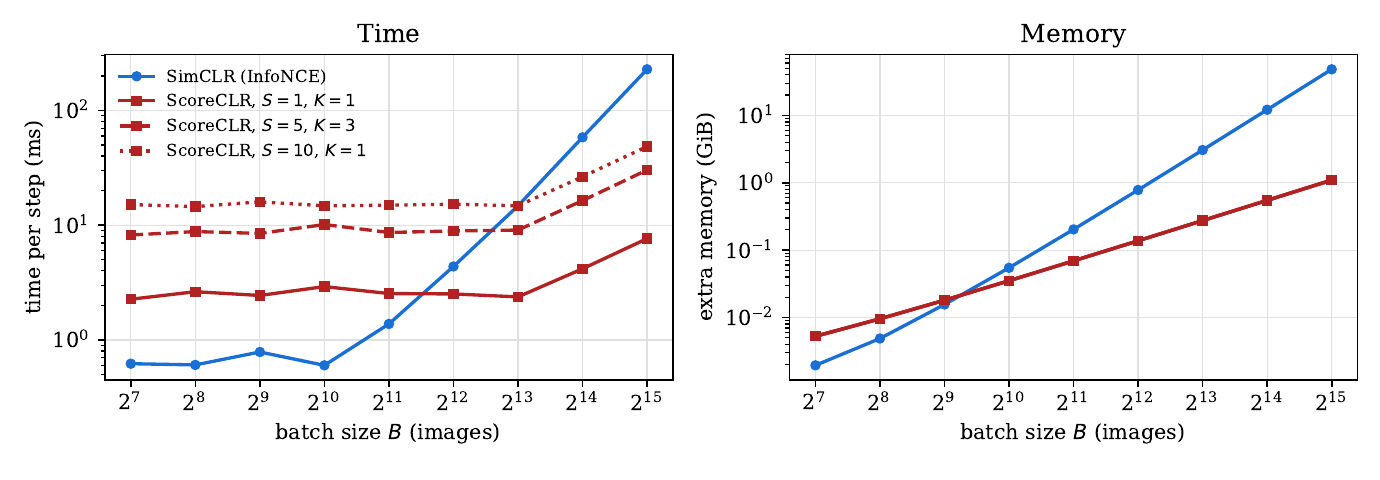}
\caption{Measured cost of the objective alone vs.\ the batch size $B$ (images), on one H100 GPU (fp32). Codes are given; we time one step of the loss (forward and backward to the codes; for ScoreCLR, $S$ score-network updates and $K$ queries, as indicated; $S{=}5$, $K{=}3$ is the ImageNet-100 recipe), median over 30 iterations, and report the extra peak memory, which does not depend on $S$ or $K$.}
\label{fig:cost}
\end{figure}

InfoNCE compares every pair of codes in the batch, so the cost of its objective grows quadratically with the batch size $B$. ScoreCLR instead passes each code through a small score network (0.28M parameters, 2.2\% of the encoder and projector) a fixed number of times, $S$ updates and $K$ queries, so the cost of its objective grows linearly with $B$. We measured both objectives on one H100 GPU in fp32 (Figure~\ref{fig:cost}, Table~\ref{tab:measuredcost}). For the objective alone, InfoNCE is faster at small batches (0.6\,ms against 2.3\,ms at $B=128$), ScoreCLR uses less memory from $B\approx1$k, and it is faster from $B\approx2$k--4k with $S=1$ and above $B\approx8$k with $S=10$, the number of score-network updates the large-batch runs need. At $B=32768$, InfoNCE takes 228\,ms and 48.3\,GiB; ScoreCLR takes 7.6\,ms with $S=1$, $K=1$, 30\,ms with the ImageNet-100 recipe ($S=5$, $K=3$) and 48.5\,ms with $S=10$, $K=1$, and 1.1\,GiB in each case. In a full training step the encoder dominates, so the gain is smaller but measured: one step of one device at $B=32768$ over 8 devices (emulated on one GPU) takes 508\,ms and 43.1\,GiB for ScoreCLR ($S=5$, $K=3$) against 531\,ms and 48.7\,GiB for SimCLR (the encoder alone takes 500\,ms; Table~\ref{tab:fullcost}), and a full epoch takes the same time within 8\% (Table~\ref{tab:epochcost}).

\begin{table}[ht]
\centering
\caption{Cost of one full training step per device (ResNet-18, CIFAR-10 resolution, fp32, one H100): $B{=}4096$ images per device, i.e.\ $32768$ images over 8 devices, emulated on one GPU. ScoreCLR: $K{=}3$.}
\label{tab:fullcost}
\small
\begin{tabular}{lcc}
\toprule
& Time (ms) & Peak memory (GiB) \\
\midrule
Encoder only (no objective) & 499.8 & 43.1 \\
ScoreCLR, $S{=}5$ & 508.2 & 43.1 \\
SimCLR & 530.9 & 48.7 \\
\bottomrule
\end{tabular}
\end{table}

\section{Conclusion}
\label{sec:conclusion}
We introduced ScoreCLR, which replaces the batch-based repulsion of contrastive learning by a score-based estimate of the entropy gradient, learned jointly with the encoder. A natural extension is a fully score-based critic, where the conditional entropy is also derived from the learned score. Since the score network predicts clean information from a noisy input, it would also be interesting to test whether the predictor of JEPA methods \citep{assran2023ijepa}, which predicts the clean latent of a target from a masked context, could by itself be a proxy to enforce high entropy. Finally, since the score-based estimator keeps the projector output at a higher effective rank than existing regularizers (Section~\ref{sec:geometry}), it would be interesting to study priors with finer geometric properties than the isotropic Gaussian and to compare with those methods in that setting. Since ScoreCLR learns the score of the latent distribution at all noise levels, we also plan on investigating whether its score network can serve as a generative model of the representation.

\subsection*{Code and reproducibility}
All training recipes, evaluation protocols and the cost model are given in Appendix~\ref{app:details}, and the proof of Proposition~\ref{prop:entropy-grad} in Appendix~\ref{app:proofs}. The ScoreCLR implementation is a self-contained PyTorch code base, and the SimCLR (LARS) baselines use the public solo-learn library \citep{dacosta2022sololearn}; the code is available at \url{https://github.com/chris-mrn/ScoreCLR_code}.

\subsection*{AI use statement}
\label{sec:aistatement}

We used generative AI tools (Claude Code) to implement the training, evaluation and cost-measurement code of ScoreCLR and of the baselines, to launch and monitor experiments, to suggest ablations, to draft the analysis of the experimental results and parts of the experimental section, to assist in writing proofs, to create figures, to edit the text for grammar, to identify relevant literature and to format references. We did not use them to generate data, to develop theoretical models, to formulate mathematical claims, to provide critical ingredients of proofs, or to propose hypotheses. The authors verified the code against the equations of the paper, checked every number in the tables against the training logs and read all suggested references. We take full responsibility for the content of this paper.

\bibliography{iclr2027_conference}
\bibliographystyle{iclr2027_conference}

\appendix

\section{Proofs}
\label{app:proofs}

\begin{assumption}
\label{ass:entropy-grad}
\leavevmode
\begin{enumerate}
\item[(i)] For every $x$, the map $\varphi \mapsto F^\varphi(x)$ is continuously differentiable.
\item[(ii)] $Z^\varphi$ has a density $p_{Z^\varphi}$ that is positive everywhere, continuously differentiable in
$(z, \varphi)$, with finite entropy, finite second moment and finite Fisher information
$\mathbb{E}\big[\|\nabla_z \log p_{Z^\varphi}(Z^\varphi)\|^2\big]$.
\item[(iii)] Derivatives with respect to $\varphi$ can be exchanged with expectations and
integrals (e.g.\ by dominated convergence).
\end{enumerate}
\end{assumption}

\begin{proof}[Proof of Proposition~\ref{prop:entropy-grad}]
Since $Z^\varphi = F^\varphi(Y)$, the law of the unconscious statistician gives
\[
H(Z^\varphi)
= -\,\mathbb{E}_{Z^\varphi}\big[\log p_{Z^\varphi}(Z^\varphi)\big]
= -\,\mathbb{E}_{Y}\big[\log p_{Z^\varphi}(F^\varphi(Y))\big] .
\]

The encoder parameters enter in two places: through the point $F^\varphi(Y)$ at which the
density is evaluated, and through the density $p_{Z^\varphi}$ itself. Differentiating with respect to $\varphi$,
\[
\nabla_{\varphi} H(Z^\varphi)
= -\,\mathbb{E}_{Y}\Big[\big(\nabla_{\varphi} F^\varphi(Y)\big)^{\!\top}
      \big(\nabla_z \log p_{Z^\varphi}\big)(F^\varphi(Y))\Big]
  -\,\mathbb{E}_{Y}\Big[\big(\nabla_{\varphi} \log p_{Z^\varphi}\big)(F^\varphi(Y))\Big] ,
\]
where in the second term $\nabla_\varphi$ acts on the parameter index of the density only,
the evaluation point being held fixed.

The second term vanishes. Indeed,
\[
\begin{aligned}
\mathbb{E}_{Y}\big[\big(\nabla_{\varphi} \log p_{Z^\varphi}\big)(F^\varphi(Y))\big]
&= \mathbb{E}_{Z^\varphi}\big[\nabla_{\varphi} \log p_{Z^\varphi}(Z^\varphi)\big]
 = \int p_{Z^\varphi}(z)\,\frac{\nabla_{\varphi} p_{Z^\varphi}(z)}{p_{Z^\varphi}(z)}\,\mathrm{d}z \\
&= \nabla_{\varphi} \int p_{Z^\varphi}(z)\,\mathrm{d}z
 = \nabla_{\varphi}(1) = 0 ,
\end{aligned}
\]
the interchange of $\nabla_\varphi$ and the integral being justified by dominated convergence.
Hence
\begin{equation}
\label{eq:entropy-grad}
\nabla_{\varphi} H(Z^\varphi)
= -\,\mathbb{E}_{Y}\Big[\big(\nabla_{\varphi} F^\varphi(Y)\big)^{\!\top}
      \big(\nabla_z \log p_{Z^\varphi}\big)(F^\varphi(Y))\Big] .
\end{equation}

It remains to identify the right-hand side as the gradient of the stated surrogate. Write
\[
\mathcal{L}(\varphi) = \tfrac{1}{2}\,\mathbb{E}_{Y}\big[\|F^\varphi(Y) - c^\varphi(Y)\|^2\big],
\qquad
c^\varphi(Y) := \operatorname{sg}\!\Big(F^\varphi(Y) + \big(\nabla_z \log p_{Z^\varphi}\big)(F^\varphi(Y))\Big).
\]
The stop-gradient has zero derivative, so $\nabla_\varphi c^\varphi(Y) = 0$ and only the first
occurrence of $F^\varphi$ is differentiated:
\[
\nabla_{\varphi} \mathcal{L}(\varphi)
= \mathbb{E}_{Y}\Big[\big(\nabla_{\varphi} F^\varphi(Y)\big)^{\!\top}
      \big(F^\varphi(Y) - c^\varphi(Y)\big)\Big] .
\]
At the point of evaluation the stop-gradient is the identity, so
$F^\varphi(Y) - c^\varphi(Y) = -\big(\nabla_z \log p_{Z^\varphi}\big)(F^\varphi(Y))$, and therefore
\[
\nabla_{\varphi} \mathcal{L}(\varphi)
= -\,\mathbb{E}_{Y}\Big[\big(\nabla_{\varphi} F^\varphi(Y)\big)^{\!\top}
      \big(\nabla_z \log p_{Z^\varphi}\big)(F^\varphi(Y))\Big]
= \nabla_{\varphi} H(Z^\varphi)
\]
by \eqref{eq:entropy-grad}.
\end{proof}

\section{Connection to other methods}

ScoreCLR was first motivated by a comparison between distribution matching distillation \citep{yin2024onestep} and generative modeling by drifting \citep{deng2026drifting}. We present below its links with SimCLR and with drifting models.

\subsection{SimCLR}
\label{app:SimCLRvsScoreCLR}
SimCLR normalizes the codes, so $\|Z_v^i\|=1$. Then $f_\tau(z,y) = \frac{1}{\tau} - \frac{1}{2\tau}\|z-y\|^2$ and $k_\tau(z,y) \propto \exp\!\big(-\|z-y\|^2/(2\tau)\big)$ is a Gaussian kernel of variance $\tau$. Up to an additive constant, the SimCLR loss \citep{chen2020simple} writes
\[
\mathcal{L}_{\text{SimCLR}} = \frac{1}{2\tau}\mathcal{L}_{\text{inv}} + \mathcal{L}_{\text{rep}}^{\text{SimCLR}}, \qquad
\mathcal{L}_{\text{rep}}^{\text{SimCLR}} = \frac{1}{2B}\sum_{i=1}^{B}\sum_{v\in\{1,2\}} \log\Big(\sum_{(j,u)\neq(i,v)} e^{-\|Z_v^i - Z_u^j\|^2/(2\tau)}\Big).
\]
The invariance term is the same as in ScoreCLR. Since $Z_v^i = G^\varphi(T_v(X^i))$, the gradient of the repulsion with respect to $\varphi$ is
\begin{align*}
\nabla_\varphi \mathcal{L}_{\text{rep}}^{\text{SimCLR}}
= \frac{1}{2B\tau}\sum_{i=1}^{B}\sum_{v}\big(\nabla_\varphi Z_v^i\big)^{\!\top}
\Bigg[
&\sum_{(j,u)\neq(i,v)} \frac{e^{-\|Z_v^i - Z_u^j\|^2/(2\tau)}}
{\sum_{(l,w)\neq(i,v)} e^{-\|Z_v^i - Z_w^l\|^2/(2\tau)}}\big(Z_u^j - Z_v^i\big) \\
&+\sum_{(j,u)\neq(i,v)} \frac{e^{-\|Z_u^j - Z_v^i\|^2/(2\tau)}}
{\sum_{(l,w)\neq(j,u)} e^{-\|Z_u^j - Z_w^l\|^2/(2\tau)}}\big(Z_u^j - Z_v^i\big)
\Bigg],
\end{align*}
whereas for ScoreCLR, with $t\sim p(t)$ and $\varepsilon\sim\mathcal{N}(0,I)$,
\[
\nabla_\varphi \mathcal{L}_{\text{rep}}^{\text{ScoreCLR}}
= \frac{1}{2B}\sum_{i=1}^{B}\sum_{v}\big(\nabla_\varphi Z_v^i\big)^{\!\top}\,
\mathbb{E}_{t,\varepsilon}\big[s_t^\theta\big(Z_v^i + t\varepsilon\big)\big].
\]
The first term of the bracket is the score of a Gaussian kernel density estimator (KDE) of bandwidth $\sqrt{\tau}$ built on the $2B-1$ other codes, i.e.\ a batch estimate of the score of $p_{Z_t^\varphi}$ at $t=\sqrt{\tau}$. It is the counterpart of the ScoreCLR term, with two differences: (i) it costs $\mathcal{O}(B^2)$ and uses a fixed $t=\sqrt{\tau}$, whereas ScoreCLR uses a learned score at a random $t\sim p(t)$ at cost $\mathcal{O}(B)$; (ii) it is evaluated at the clean code $Z_v^i$, whereas ScoreCLR evaluates it at $Z_v^i + t\varepsilon$.

The second term has no counterpart in ScoreCLR. It appears because $Z_v^i$ is also a kernel centre in the sums of the other anchors, so moving $Z_v^i$ changes the estimated density at the other codes. For the true density this effect is zero, as in the proof of Proposition~\ref{prop:entropy-grad}:
\[
\mathbb{E}\big[\nabla_\varphi \log p_{Z^\varphi}(Z)\big|_{Z=Z^\varphi}\big]
= \int \nabla_\varphi p_{Z^\varphi}(z)\,\mathrm{d}z = \nabla_\varphi 1 = 0 .
\]
ScoreCLR never computes it, since the score network is a stop-gradient. In SimCLR, the KDE is built from the codes being differentiated, so the term is not exactly zero on a finite batch.

Finally, SimCLR has no regularization term $\mathcal{L}_{\text{reg}}$: the constraint $\|Z_v^i\|=1$ plays the role of the Gaussian prior.

\subsection{Drifting models}
\label{app:drifting}

Considering the terminology of \citet{deng2026drifting}, we denote as drifting field
\[
\mathrm{drift}_i^\varphi = -\nabla_z \log p_{Z^{\varphi}}(Z_i^{\varphi}) - Z_i^{\varphi}, \qquad i \in \{1,2\},
\]
the velocity field of the Wasserstein gradient flow of $KL(\cdot\,\|\,\mathcal{N}(0,I))$ \citep{gretton2026wasserstein}. Symmetrizing the repulsion force with respect to the views and using the stop-gradient trick of DMD \citep{yin2024onestep}, we get the mutual information loss with Gaussian prior, which writes as
\[
\mathcal{L}_{\text{MI-score}}^{\text{Gaussian prior}}(\varphi) = \lambda\,\mathbb{E}\big[\|Z_1^\varphi-Z_2^\varphi\|^2\big] + \frac{1}{2}\sum_{i=1}^{2}\mathbb{E}\big[\|Z_i^\varphi - \text{sg}(Z_i^\varphi + \mathrm{drift}_i^\varphi)\|^2\big],
\]
with the same convention as Eq.~(\ref{eq:scoreclr_loss}): $\lambda$ weights the invariance term and the prior term has weight one.

\section{Experimental Details}
\label{app:details}

\subsection{Pre-training protocol}
\label{app:protocol}

\paragraph{Shared recipe.}
SimCLR, LeJEPA, BYOL, Barlow Twins and ScoreCLR share the ResNet-18 encoder, the two-layer MLP projector (Linear--ReLU--Linear, 2048 hidden units, 256 outputs, no BatchNorm), the optimizer (AdamW, weight decay $10^{-6}$), the gradient-norm clip $1.0$, the schedule (constant learning rate, then a cosine decay to zero over the last 100 epochs), the batch size ($128$ for 1000 epochs on CIFAR-10, $2\times128$ for 400 epochs on ImageNet-100), the augmentations and fp32 precision (Table~\ref{tab:hparams}). The clip applies to the encoder, the projector and the online classifier and, for ScoreCLR, also to the score network. Only the encoder learning rate is tuned per method: $2{\times}10^{-4}$ for SimCLR, LeJEPA, BYOL and Barlow Twins, and $10^{-4}$ (CIFAR-10) or $2{\times}10^{-4}$ (ImageNet-100) for ScoreCLR.

\paragraph{Augmentations.}
SimCLR, LeJEPA, BYOL, Barlow Twins and ScoreCLR use the same symmetric pipeline on both datasets: random resized crop (scale $[0.08,1]$), horizontal flip (probability $0.5$), colour jitter $(0.8,0.8,0.8,0.2)$ applied with probability $0.8$, grayscale with probability $0.2$ and Gaussian blur with probability $0.5$, without solarization. The SimCLR (LARS) rows come from solo-learn's own pipeline (data loading and augmentations) and are reference points, not controlled comparisons.

\paragraph{Baselines.}
\emph{SimCLR} uses the symmetrized InfoNCE loss with temperature $\tau=0.2$ and negatives gathered across devices. \emph{LeJEPA} replaces the repulsion by SIGReg ($\lambda=0.02$, 17 knots, $t_{\max}=3$, 256 slices). \emph{BYOL} uses a 4096-unit predictor without BatchNorm and an EMA coefficient that follows a cosine from $0.99$ to $1$; \emph{Barlow Twins} uses $\lambda_{\mathrm{BT}}=0.0051$ and a loss scale of $0.1$, both solo-learn's CIFAR-10 values. BYOL and Barlow Twins are run in this harness with seeds 1--3; removing BatchNorm from their predictor and projector disadvantages them (BYOL reaches $89.96$\% linear accuracy with a BatchNorm predictor, seed 1). The \emph{SimCLR (LARS)} rows follow solo-learn's recipe: LARS (momentum $0.9$, $\eta=0.02$, biases and normalization layers excluded), weight decay $10^{-4}$, a 10-epoch warm-up from $3{\times}10^{-5}$ followed by a cosine decay, $\tau=0.2$, 16-bit mixed precision, learning rate $0.4$ with batch $256$ on CIFAR-10 (1000 epochs) and learning rate $0.3$ with batch $2\times128$ on ImageNet-100 (400 epochs). The projector is narrowed from the published $4096/512$ to the shared $2048/256$ head, except in the control row, which keeps the published head and every other published default.

\subsection{ScoreCLR implementation}
\label{app:scoreclr}

\paragraph{Loss.}
We minimize Eq.~(\ref{eq:scoreclr_loss}), $\mathcal{L}=\lambda\,\mathcal{L}_{\text{inv}}+\mathcal{L}_{\text{rep}}+\mathcal{L}_{\text{reg}}$, with $\lambda=5$ on CIFAR-10 and $\lambda=2$ on ImageNet-100. In the Gaussian conditional model used to derive the loss, $\lambda=1/(2\sigma^2)$, i.e.\ $\sigma^2=0.1$ and $0.25$. The repulsion and the prior carry weight one: they are two parts of one estimate of $\mathrm{KL}(p_Z\,\|\,\mathcal{N}(0,I))$ and are not tuned separately. Each of the three terms is divided by the code dimension $d=256$; this common factor leaves the relative weights unchanged and only matters through the gradient clip. The repulsion and the prior are averaged over $K$ independent noise queries per code, with $K=1$ on CIFAR-10 and $K=3$ on ImageNet-100. The noise levels follow $\ln t\sim\mathcal{N}(-1.2,\,p_{\mathrm{std}}^2)$ with $p_{\mathrm{std}}=1.2$ on CIFAR-10 and $1.0$ on ImageNet-100.

\paragraph{Score network.}
The score network is an unconditional noise-prediction MLP $\varepsilon^\theta(z_t,t)$: three linear layers of width $256$ with SiLU activations on the concatenation of $z_t$ and a 64-dimensional sinusoidal embedding of $\ln t$, the two hidden layers being FiLM-modulated by the same embedding (modulation initialized to the identity). It has $0.28$M parameters, 2.2\% of the encoder and projector. It is trained with the unweighted loss $\mathcal{L}_{\text{DSM}}$ on the detached codes of both views, with fresh noise for each of the $S$ updates per encoder step ($S=1$ on CIFAR-10, $S=5$ on ImageNet-100), and queried with $s_t^\theta=-\varepsilon^\theta/t$. Its optimizer is AdamW with learning rate $10^{-3}$, weight decay $10^{-2}$, gradient clip $1.0$. Like the encoder's, its learning rate is constant and then follows a cosine decay to zero over the last 100 epochs.

\begin{table}[ht]
\centering
\caption{Pre-training hyperparameters (CIFAR-10 / ImageNet-100 where two values are given). ``Shared AdamW'' is the recipe of SimCLR, LeJEPA, BYOL and Barlow Twins. ImageNet-100 uses $2\times128$ images per step with DDP-averaged gradients and no learning-rate rescaling.}
\label{tab:hparams}
\scriptsize
\setlength{\tabcolsep}{3pt}
\begin{tabular}{lp{3.1cm}p{3.1cm}p{4.3cm}}
\toprule
& Shared AdamW & SimCLR (LARS) & ScoreCLR \\
\midrule
Backbone; epochs & ResNet-18; 1000 / 400 & ResNet-18; 1000 / 400 & ResNet-18; 1000 / 400 \\
Batch size & 128 / $2{\times}128$ & 256 / $2{\times}128$ & 128 / $2{\times}128$ \\
Optimizer & AdamW & LARS & AdamW (encoder and score net) \\
Encoder LR & $2{\times}10^{-4}$ & 0.4 / 0.3 & $10^{-4}$ / $2{\times}10^{-4}$ \\
Weight decay & $10^{-6}$ & $10^{-4}$ & $10^{-6}$ (score net $10^{-2}$) \\
LR schedule & constant, cosine over last 100 ep & 10-ep warm-up + cosine & constant, cosine over last 100 ep (encoder and score net) \\
Gradient clip & 1.0 & none & 1.0 (encoder and score net) \\
Precision & fp32 & 16-bit mixed & fp32 \\
Augmentations & symmetric pipeline (Appendix~\ref{app:protocol}) & solo-learn's own & same as shared AdamW \\
Projector & \multicolumn{3}{c}{Linear--ReLU--Linear, no BatchNorm, 2048 / 256 (LARS control: 4096 / 512)} \\
Classifier LR & 0.01 & 0.1 & 0.01 \\
Objective & method-specific & $\tau=0.2$ & $\lambda=5$ / $2$; repulsion and prior weight 1 \\
Score network & -- & -- & unconditional $\varepsilon$-MLP-FiLM, $3{\times}256$, 64-d emb., 0.28M \\
Score-net LR; steps $S$ & -- & -- & $10^{-3}$ ($10\times$ / $5\times$ encoder); 1 / 5 \\
Queries $K$ & -- & -- & 1 / 3 \\
DSM noise $\ln t$ & -- & -- & $\mathcal{N}(-1.2,\,1.2^2)$ / $\mathcal{N}(-1.2,\,1.0^2)$ \\
\bottomrule
\end{tabular}
\end{table}

\subsection{Linear evaluation}
\label{app:probe}

The offline probe trains a linear classifier for 100 epochs on frozen backbone features in fp32, with the standard crop-and-flip augmentation, using the same protocol for every method; the only exception is the $k$-NN accuracy of the SimCLR (LARS) rows, computed with a memory bank under crop-and-flip augmentation. On CIFAR-10 the features are used raw and the classifier is trained with SGD (momentum 0.9, learning rate 0.1, divided by 10 at epochs 60 and 80). On ImageNet-100 the features are L2-normalized and the classifier is trained with SGD (momentum 0.9, learning rate 10, warm-up then cosine). The $k$-NN accuracy is a weighted cosine $k$-NN ($k=20$, temperature $0.07$) of the validation features against the clean training set. Alignment is computed on L2-normalized projector outputs, uniformity on 4000 validation codes, and RankMe on the 10{,}000 validation codes.

\subsection{Effective rank}
\label{app:rank}

RankMe \citep{garrido2023rankme} is the exponential of the entropy of the
normalized singular values of the feature matrix, computed on the validation
set. We report it on both the backbone features and the projector output
because \citet{jing2022understanding} (DirectCLR) showed that contrastive
learning can collapse the projector output (dimensional collapse) while the backbone stays full-rank.

\subsection{Out-of-domain transfer}
\label{app:transfer}

Following \citet{kornblith2019better}, a multinomial logistic-regression
classifier is fit on frozen, L2-normalized backbone features of each
downstream training set. The L2 penalty is selected on a validation split
(the dataset's own where one exists, else a stratified 20\% hold-out) and the
classifier is refit on train\,+\,val. Downstream images are resized and
center-cropped to the pretraining resolution: $32{\times}32$ with the CIFAR
stem for CIFAR-10 backbones and $224{\times}224$ for ImageNet-100 backbones.

\subsection{Ablations of the recipe}
\label{app:ablations}

Table~\ref{tab:ablations} varies ingredients of the CIFAR-10 recipe (batch 128, 1000 epochs, $\lambda=5$). The reference is $K{=}1$, $S{=}1$, $p_{\mathrm{std}}{=}1.2$ with clip $1.0$ (mean $\pm$ standard deviation over seeds 1--3); a paired row is compared with the reference run of the same seed.

\begin{table}[ht]
\centering
\caption{Ablations on CIFAR-10: linear accuracy (\%), $k$-NN accuracy (\%) and RankMe of $z$. ``Ref.'' is the linear accuracy of the same-seed run with the same noise width and clip. $^{\dagger}$: seeds 5 and 42, run with an earlier code snapshot that implements the same loss. The $K{=}3$ row uses seeds 1--3 without gradient clipping.}
\label{tab:ablations}
\small
\setlength{\tabcolsep}{3.5pt}
\begin{tabular}{llcccc}
\toprule
Variant & Seeds & Linear & $k$-NN & RankMe $z$ & Ref. \\
\midrule
Reference ($K{=}1$, $S{=}1$, $p_{\mathrm{std}}{=}1.2$, clip) & 1--3 & $88.87{\pm}0.44$ & $88.27{\pm}0.13$ & 229 & -- \\
$p_{\mathrm{std}}{=}1.0$ & 1--3 & $88.57{\pm}0.19$ & $88.20{\pm}0.08$ & 233 & -- \\
Conditional score network & 1 / 2 & 83.83 / 85.27 & 83.54 / 84.34 & 25 / 51 & 89.38 / 88.58 \\
$K{=}1$, no clipping$^{\dagger}$ & 5, 42 & 88.87 / 88.85 & 88.44 / 88.43 & 229 & -- \\
$K{=}3$, no clipping & 1--3 & $78.62{\pm}4.97$ & $78.94{\pm}4.51$ & $105{\pm}32$ & -- \\
\bottomrule
\end{tabular}
\end{table}

\paragraph{Conditioning.} A score network that also receives a second latent as input, namely the other view of the same image when it is trained to denoise and the code of another image of the batch when it is queried, loses 3.3 to 5.6 points of linear accuracy with respect to the unconditional network (Table~\ref{tab:ablations}). The recipe was not re-tuned for the conditional network, so this ablation measures the conditioning inside the present recipe only.

\paragraph{Queries and clipping.} Averaging $K{=}3$ queries without gradient clipping is unstable across seeds (79.45, 83.16 and 73.26 for seeds 1--3), whereas $K{=}1$ without clipping gives 88.87 and 88.85. We use the clip for fairness with the baselines, for which it is neutral (SimCLR: 89.41 without it, 89.45 with it, seed 5). On ImageNet-100, $K{=}3$ with the clip is stable ($71.23{\pm}0.06$ over three seeds).

\paragraph{Score-network steps.} Extra score-network steps matter when iterations are scarce. On ImageNet-100 ($494$ iterations per epoch), $K{=}1$ and $S{=}1$ reach $57.14$ against $71.23{\pm}0.06$ for the recipe of Table~\ref{tab:hparams}. At batch 2048 on CIFAR-10 ($24$ iterations per epoch), $S{=}1$ is not sufficient, $S{=}5$ gives 87.84 and $S{=}10$ 88.29.

\paragraph{Larger batches.} At $B{=}4096$, $S{=}10$ gives 87.50 with $p_{\mathrm{std}}{=}1.2$ (Table~\ref{tab:batch}) and 87.56 with $p_{\mathrm{std}}{=}1.0$; with $p_{\mathrm{std}}{=}1.0$, $S{=}5$ gives 85.76. 

\subsection{Cost model}
\label{app:cost}

With $N=2B$ codes of dimension $d=256$, InfoNCE's similarity matrix costs
$6N^2d$ FLOPs and about $16N^2$ bytes. ScoreCLR queries its score network
($P=0.28$M parameters) separately for each code with $S=5$ score-matching
updates and $K=3$ repulsion queries, about $2P(6S+2K)=20$\,MFLOPs per image
(6.7\,MFLOPs with $S=1$). The CIFAR-10 recipe ($S=1$, $K=1$) costs $2P(6+2)=4.5$\,MFLOPs per image; Tables~\ref{tab:fullcost} and~\ref{tab:cost} use the heavier ImageNet-100 setting; Figure~\ref{fig:cost} and Table~\ref{tab:measuredcost} report three settings. In fp32, one score-matching step saves 30.2\,KB per image for backward; the $S$ steps are sequential and the repulsion queries run without gradient, so this does not depend on $S$ and $K$. The loss saves 9.2\,KB more, which gives 39\,KB per image. Architecture constants are counted on the implemented
modules. For the full-iteration memory ratio, activations saved for backward
are counted in 16-bit mixed precision; under data parallelism over $W$ devices
each device holds a $2B/W\times2B$ block of the similarity matrix, so the
ratio is independent of $W$. At $B=32768$ ScoreCLR's objective needs
$10\times$ fewer FLOPs and $53\times$ less memory than InfoNCE, and the
per-device saving is 27\% for ResNet-18 at 32\,px and 8\% at 224\,px;
SimCLR's objective alone exceeds one 80\,GB GPU at $B=65536$.

\begin{table}[ht]
\centering
\caption{Analytical cost vs.\ global batch size $B$ (ScoreCLR: $S{=}5$, $K{=}3$). Left: the objective alone. Right: ScoreCLR / SimCLR per-device memory for a full data-parallel iteration.}
\label{tab:cost}
\small
\setlength{\tabcolsep}{3.5pt}
\begin{tabular}{rcccccc}
\toprule
& \multicolumn{4}{c}{Objective only} & \multicolumn{2}{c}{Full iteration, memory ratio} \\
\cmidrule(lr){2-5} \cmidrule(lr){6-7}
& \multicolumn{2}{c}{TFLOPs} & \multicolumn{2}{c}{Memory (GB)} & R18 & R18 \\
$B$ & SimCLR & ScoreCLR & SimCLR & ScoreCLR & 32\,px & 224\,px \\
\midrule
4096  & 0.10 & 0.08 & 1.1  & 0.16 & 0.96 & 0.99 \\
8192  & 0.41 & 0.17 & 4.3  & 0.32 & 0.92 & 0.98 \\
16384 & 1.65 & 0.33 & 17.2 & 0.65 & 0.85 & 0.96 \\
32768 & 6.60 & 0.66 & 68.7 & 1.29 & 0.73 & 0.92 \\
65536 & 26.4 & 1.32 & 275  & 2.58 & 0.57 & 0.85 \\
\bottomrule
\end{tabular}
\end{table}

\paragraph{Measured cost.}
Table~\ref{tab:measuredcost} gives the measured time and extra peak memory of the objective alone, in the protocol of Figure~\ref{fig:cost}, and Table~\ref{tab:epochcost} the time of a full training epoch, measured for both methods in the same job on the same GPU. In a full epoch the encoder dominates: ScoreCLR takes 1.08 times the time of SimCLR on CIFAR-10 at $B=128$, 1.01 times on ImageNet-100, and the same time within 3\% from $B=2048$ to $8192$, where it uses 2\% less memory at $B=4096$ and 6\% less at $B=8192$ in 16-bit precision (the model of Table~\ref{tab:cost} gives 4\% and 8\%). At $B=32768$ the measured objective uses 44 times less memory than InfoNCE (53 in the model), and the per-device saving of one step measured in fp32 is 11\% (Table~\ref{tab:fullcost}), against 27\% in the model, which counts 16-bit activations.

\begin{table}[ht]
\centering
\caption{Measured cost of the objective alone (one H100, fp32, median over 30 iterations; same job as Figure~\ref{fig:cost}): time per step and extra peak memory. The memory of ScoreCLR depends on neither $S$ nor $K$.}
\label{tab:measuredcost}
\small
\setlength{\tabcolsep}{4pt}
\begin{tabular}{rcccccc}
\toprule
& \multicolumn{4}{c}{Time per step (ms)} & \multicolumn{2}{c}{Extra memory} \\
\cmidrule(lr){2-5} \cmidrule(lr){6-7}
& & \multicolumn{3}{c}{ScoreCLR} & & \\
\cmidrule(lr){3-5}
$B$ & InfoNCE & $S{=}1$, $K{=}1$ & $S{=}5$, $K{=}3$ & $S{=}10$, $K{=}1$ & InfoNCE & ScoreCLR \\
\midrule
128   & 0.62 & 2.3 & 8.2  & 15.1 & 2.0\,MiB   & 5.4\,MiB \\
512   & 0.78 & 2.4 & 8.5  & 16.0 & 16\,MiB    & 19\,MiB \\
2048  & 1.4  & 2.5 & 8.6  & 14.9 & 208\,MiB   & 71\,MiB \\
4096  & 4.4  & 2.5 & 8.9  & 15.2 & 800\,MiB   & 141\,MiB \\
8192  & 14.7 & 2.4 & 9.1  & 14.8 & 3.1\,GiB   & 281\,MiB \\
32768 & 228  & 7.6 & 30.3 & 48.5 & 48.3\,GiB  & 1.1\,GiB \\
\bottomrule
\end{tabular}
\end{table}

\begin{table}[ht]
\centering
\caption{Measured time of one training epoch and peak memory, ScoreCLR / SimCLR (CIFAR-10: one H100; ImageNet-100: two H100s, memory per GPU). ``--'': not recorded. The $B{=}128$ row was timed with $K{=}3$ queries, more work per step than the $K{=}1$ recipe of Table~\ref{tab:linear}.}
\label{tab:epochcost}
\small
\begin{tabular}{llcc}
\toprule
Setting & ScoreCLR & Epoch time (s) & Peak memory (GiB) \\
\midrule
CIFAR-10, $B{=}128$, fp32 & $S{=}1$, $K{=}3$ & 8.96 / 8.28 & -- \\
ImageNet-100, $B{=}2{\times}128$, fp32 & $S{=}5$, $K{=}3$ & 71.0 / 70.3 & 5.63 / 5.62 \\
CIFAR-10, $B{=}2048$, fp16 & $S{=}10$, $K{=}1$ & 6.81 / 6.64 & 11.0 / 11.1 \\
CIFAR-10, $B{=}4096$, fp16 & $S{=}10$, $K{=}1$ & 8.27 / 8.28 & 21.7 / 22.2 \\
CIFAR-10, $B{=}8192$, fp16 & $S{=}10$, $K{=}1$ & 10.00 / 9.98 & 43.1 / 45.8 \\
\bottomrule
\end{tabular}
\end{table}

\paragraph{Overhead at small batch.} The time of the ScoreCLR objective is nearly flat from $B=128$ to $8192$ ($2.3$--$2.9$\,ms with $S=1$, $K=1$; Figure~\ref{fig:cost}, Table~\ref{tab:measuredcost}) while the work grows by a factor 64, so at small batch the time is a fixed cost, set by the number of small operations in the score-network step (three linear layers, the modulation, the gradient clipping and the optimizer step) rather than by the amount of computation. We did not optimize this. A fused kernel for the score network and its update, or capturing the step in a CUDA graph, should reduce this fixed cost; we have not tested it.

\subsection{Estimator of the Poincar\'e constant}
\label{app:poincare}

The Poincar\'e constant is $C_P=\sup_f \mathrm{Var}_p(f)/\mathbb{E}_p\|\nabla f\|^2$, and we estimate it by maximizing this ratio over a class of test functions. Following \citet{pillaudvivien2020statistical}, we take $f$ in the span of the coordinates of $z$ and of $M=1024$ random Fourier features $\cos(w_k^\top z+b_k)$, with $w_k\sim\mathcal{N}(0,(\ell m)^{-2}I)$, $b_k$ uniform on $[0,2\pi]$ and $m$ the median distance between codes. The ratio is then a generalized Rayleigh quotient between the covariance of the features and the Gram matrix of their gradients, and it is maximized by a generalized eigenvector. We fit it on half of the validation codes, select the bandwidth factor $\ell\in\{0.25,0.5,1,2\}$, a ridge term and the eigenvector on a quarter, and report the ratio on the last quarter, because the ratio measured on the fitting codes overestimates the constant. The constant scales as the square of the codes, so we first center and whiten them (PCA, keeping the directions whose variance exceeds $10^{-6}$ times the largest; 174 of 256 directions for SimCLR, whose output is collapsed). The estimate then reflects the shape of the distribution rather than its scale or anisotropy, and it equals 1 for a Gaussian. Since the estimate uses a fixed class of test functions, a smaller value is an indication, not a proof, that the codes are closer to a Gaussian.

\end{document}